# A consumer BCI for Automated Music Evaluation within a popular on-demand Music Streaming Service

''Taking Listener's Brainwaves to Extremes''

Fotis Kalaganis[1], Dimitrios A. Adamos[2,3], Nikos Laskaris[1,3]

[1]AIIA Lab, Department of Informatics,
[2]School of Music Studies
[3]Neuroinformatics GRoup, http://neuroinformatics.gr
Aristotle University of Thessaloniki, 54124 Thessaloniki, Greece
kalaganis@csd.auth.gr
dadam@mus.auth.gr, d.adamos@ieee.org
laskaris@aiia.csd.auth.gr

**Abstract.** We investigated the possibility of using a machine-learning scheme in conjunction with commercial wearable EEG-devices for translating listener's subjective experience of music into scores that can be used for the automated annotation of music in popular on-demand streaming services.

Based on the established -neuroscientifically sound- concepts of brainwave frequency bands, activation asymmetry index and cross-frequency-coupling (CFC), we introduce a Brain Computer Interface (BCI) system that automatically assigns a rating score to the listened song.

Our research operated in two distinct stages: i) a generic feature engineering stage, in which features from signal-analytics were ranked and selected based on their ability to associate music induced perturbations in brainwaves with listener's appraisal of music. ii) a personalization stage, during which the efficiency of extreme learning machines (ELMs) is exploited so as to translate the derived patterns into a listener's score. Encouraging experimental results, from a pragmatic use of the system, are presented.

## 1 Introduction

Until recently, electroencephalography (EEG) was exclusive to doctors' facilities and specialists' workplaces, where trained experts operated expensive devices. The vast majority of the research was dedicated to the diagnosis of epilepsy, sleep disorders, Alzheimer's disease as well as monitoring certain clinical procedures such as anesthesia. By the same token, the application of Brain-Computer Interfaces (BCIs) has so far been confined to neuroprosthetics and for building communication channels for the physically impaired people [1].

Recent advances in medical sensors offer commercial EEG headsets at affordable prices. The procedure of recording the electrical activity of the brain via electrodes on the human scalp is pretty simple nowadays and does not require any expertise [2]. The latter can lead to a wide variety of commercial applications in almost any aspect of everyday life. Meanwhile, many researchers are drawn into the exploration of brain patterns that are not related to healthcare [3]. At the same time, the manufacturing of new portable neuroimaging devices favors innovation, as novel applications of non-invasive BCIs are anticipated within real-life environments [4].

Since the beginning of the twenty-first century, the digital revolution has radically affected the music industry and is continuously reforming the business model of music economy [5]. Until now, previously-established channels of music distribution have been replaced and new industry stakeholders have emerged. Among them, on-demand music recommendation and streaming services emerge as the ''*disruptive innovators*'' [6] of the new digital music ecosystem.

In our previous work [7], we presented our vision for the integration of bio-personalized features of musical aesthetic appreciation into modern music recommendation systems to enhance user's feedback and rating processes. With the current work, we attempted the first step toward implementation with the realization of an automated music evaluation process, that is performed in nearly real-time by decoding aesthetic brain responses during music listening and feeding them back to a contemporary music streaming service (i.e. Spotify) as listener's ratings. We aimed for a flexible BCI system that could easily adapt to new users and, after a brief training session, would reliably predict the listener's ratings about the listened songs.

To facilitate convenience and friendliness of the user's experience, the recording of brain activity was implemented using a modern commercial wireless EEG device. Thus, we first adapted our approach to the device capabilities, investigating which combinations of available brainwave descriptors and electrode sites are reliably and consistently reflecting the listeners' evaluation about the music being played. Then we synthesized these descriptors into a composite biomarker, common for all listeners. Finally, we examined different learning machines that could incorporate this biomarker and generalize efficiently from a very small set of training examples (paired biomarker-patterns and ratings).

The proposed music-evaluation BCI relies on standard ELMs for translating a particular set of readily-computable signal descriptors, as extracted from our 4-channel wearable EEG device, into a single numbered score expressing the appraisal of music within the range [1-5]. It can be personalized with very limited amount of training and runs with negligible amount of delay. In all our experimentations the passive listening paradigm was followed, since this is closer to the real life situation where someone enjoys listening to music.

The preliminary results, reported in this paper, include evidence about the existence of a robust set of brain activity characteristics that reliably reflect a listener's appraisal. Moreover, the effectiveness of ELMs in the particular regression task is established, by comparing its performance with alternative learning machines. Overall, the outcomes of this work are very encouraging for conducting experiments about music perception in real-life situations and embedding brain signal analysis within the contemporary technological universe.

The remaining paper is structured as follows. Section 2 serves as an introduction to EEG and its role in describing and understanding the effects of music. Section 3 outlines the essential tools that were employed during data analysis. Section 4 describes the experimental setup and the adopted methodology for analyzing EEG data. Section 5 is devoted to the presentation of results, while the last section includes a short discussion about the limitations of this study and its future perspectives.

## 2    Electroencephalography and Music perception studies

Electroencephalogram is a recording of the electrical changes occurring in the brain, produced by placing electrodes on the scalp and monitoring the developed electrical fields. EEG reflects mainly the summation of excitatory and inhibitory postsynaptic potentials at the dendrites of ensembles of neurons with parallel geometric orientation. While the electrical field produced by distinct neurons is too weak to be recorded with surface EEG electrode, as neural action gets to be synchronous crosswise a huge number of neurons, the electrical fields created by individual neurons aggregate, resulting to effects measurable outside the skull [8].

The EEG brain signals, also known as *brainwaves*, are traditionally decomposed (by means of band-pass filtering or a suitable transform) and examined within particular frequency bands, which are denoted via greek letters and in order of increasing central frequency are defined as follows: $\delta$ (0.5-4)Hz, $\theta$ (4-8)Hz, $\alpha$ (8-13)Hz, $\beta$ (13-30)Hz, $\gamma$(>30)Hz. EEG is widely recognized as an invaluable neuroimaging technique with high temporal resolution. Considering the dynamic nature of music, EEG appears as the ideal technique to study the interaction of music, as a continuously delivered auditory stimulus, with the brainwaves. For more than two decades neuroscientists study the relationship between listening to music and brain activity from the perspective of induced emotions [9,10,11]. More recently, a few studies appeared which shared the goal of uncovering patterns, lurked in brainwaves, that correspond to subjective aesthetic pleasure caused by music [12,13].

Regarding music perception, the literature has reported a wide spectrum of changes in the ongoing brain activity. This includes a significant increase of power in $\beta$-band over posterior brain regions [14]. An increase in $\gamma$ band, which was confined to subjects with musical training [15], an asymmetrical activation pattern reflecting induced emotions [16] and an increase of frontal midline $\theta$ power when contrasting pleasant with unpleasant musical sounds [17].

Regarding the particular task of decoding the subjective evaluation of music from the recorded brain activity, the role of higher-frequency brainwaves has been identified [18], and in particular the importance of $\gamma$-band brainwaves recorded over forebrain has been reported [19]. More recently, a relevant CFC biomarker based on the concept of nested oscillations in the brain was introduced for the assessment of spontaneous aesthetic brain responses during music listening [7]. The reported experimental results indicated that the interactions between $\beta$ and $\gamma$ oscillations, as reflected in the brainwaves recorded over the left prefrontal cortex, are crucial for estimating the subjective aesthetic appreciation of a piece of music.

## 3  Methods

This section presents briefly the methodological elements employed in the realization of the proposed framework. More specifically, instantaneous signal energy, activation asymmetry index and a CFC estimator were used to derive brainwave descriptors (i.e. features reflecting neural activity associated with music listening). The importance of each descriptor (or combination of descriptors) was evaluated using *Distance Correlation*. ELMs, an important class of artificial neural network (ANNs), were exploited to convert the derived patterns into scores that represented the listener's appraisal.

### 3.1  Signal Descriptors

Brainwaves are often characterized by their prominent frequency and their (signal) energy content. Here, we adopted a quasi-instantaneous parameterization of Brainwaves content, by means of Hilbert transform. The signal from each sensor x(t), was first filtered within the range corresponding to the frequency-band of a brain rhythm (like δ-rhythm) and the envelope of the filtered activity $α_{rhythm}(t)$ was considered as representing the momentary strength of the associated oscillatory activity. Apart from the amplitude of each brain rhythm, its relative contribution was also derived by normalizing with the total signal strength (summed from all brain rhythms).

In neuroscience research, activation refers to the change in EEG activity in response to a stimulus and is of great interest to investigate differences in the way the two hemispheres are activated [20]. To this end, an activation asymmetry index was formed by combining measurements of activation strength from two symmetrically located sensors, i.e. $AI(t) = {}^{left}α_{rhythm}(t) - {}^{right}α_{rhythm}(t)$. The normalized version of this index was also employed as an additional alternative descriptor.

A third brainwaves' descriptor was based on the CFC concept, which refers to the functional interactions between distinct brain rhythms [21]. A particular estimator was employed [22] that quantified the dependence of amplitude variations of a high-frequency brain rhythm on the instantaneous phase of a lower-frequency rhythm (a phenomenon known as phase-amplitude coupling (PAC)). This estimator operated on each sensor separately and used to investigate all the possible PAC couplings among the defined brain rhythms.

It is important to notice here, that the included descriptors were selected so as to cover different neural mechanisms and share a common algorithmic framework. Their implementation -and mainly their integration- within a unifying system did not induce time delays unreasonable for the purposes of our real-time application.

### 3.2  Distance Correlation

In statistics and in probability theory, distance correlation is a measure of statistical dependence between two random variables or two random vectors of arbitrary, not necessarily equal, dimension. Distance Correlation, denoted by $\mathcal{R}$, generalizes the idea of correlation and holds the important property that $\mathcal{R}(X,Y)=0$ if and only if X and Y are independent. $\mathcal{R}$ index satisfies $0 \leq \mathcal{R} \leq 1$ and, contrary to Pearson's correlation coefficient, is suitable for revealing non-linear relationships [23]. In this work, it was the core mechanism for identifying neural correlates of subjective music evaluation, by detecting associations between the results of signal descriptors (or combinations of them) and the listener's scores.

### 3.3  Learning Machines

Machine learning deals with the development and implementation of algorithms that can build models able to generalize a particular function from a set of given examples. Regression is a supervised learning task that machine learning can handle with efficiency of similar, or even higher, level than the standard statistical techniques, and -mainly- without imposing hypotheses. In this work, the decoding of subjective music evaluation was cast as a (nonlinear) regression problem. A model was then sought (i.e. learned from the experimental data) that would perform the mapping of patterns derived from the brain activity descriptors to the subjective music evaluation.

ELMs appeared as a suitable choice due to their documented ability to handle efficiently difficult tasks without demanding extensive training sessions [24]. They are feedforward ANNs with a single layer of hidden nodes, where the weights connecting inputs to hidden nodes are randomly assigned and never updated [25].

## 4    Implementation and Experiments

Our experimentations evolved in two different directions. First, a set of experiments were run so as to use the experimental data for establishing the brainwave pattern that would best reflect the music appraisal of an individual and train the learning machine from music pieces of known subjective evaluation. Next, additional experiments were run that implemented the real-time scoring by the trained ELM-machine so as to justify the proposed BCI in a more naturalistic setting. In both cases, the musical pieces were delivered through a popular on-demand music streaming service (Spotify) that facilitates the registration of the listener's feedback ('like'-'dislike') to adapt the musical content to his taste and make suggestions about new titles.

### 4.1    Participants

All 5 participants were healthy students of AUTH Informatics department. Their average age was 23 years and music listening was among their daily habits. They signed an informed consent after the experimental procedures had been explained to them.

### 4.2    Data Acquisition

Having in mind the user-friendliness of the proposed scheme, we adopted a modern commercial dry-sensor wireless device (i.e. Interaxon's Muse device - http://www.choosemuse.com) in our implementations. This "gadget" offers a 4-channel EEG signal, with a topological arrangement that can be seen in Fig1. The signals are digitized at the sampling frequency of 220 Hz. Data are transmitted under OSC, which is a protocol for communication among computers, sound synthesizers, and other multimedia devices that is optimized for modern networking technology [26].

**Fig. 1.** Topological arrangement of Electrodes (A) and Interaxon's Muse headset (B).

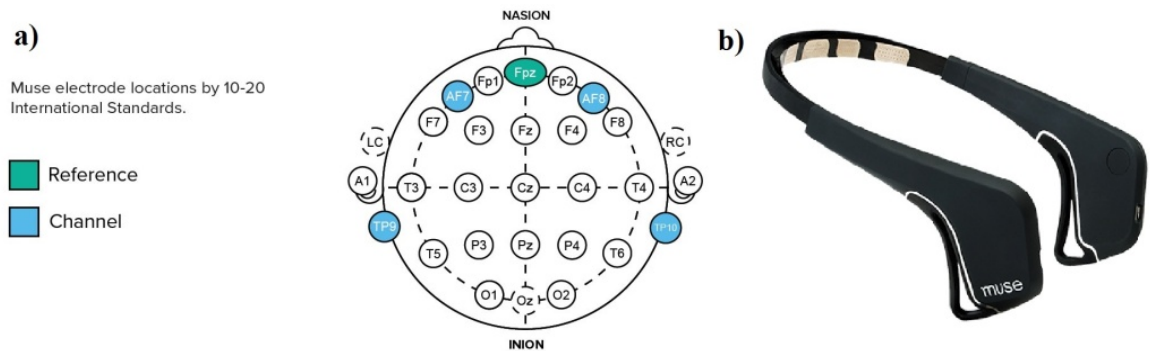

### 4.3    Experimental Procedure

Prior to placing the headset, subjects sat in a comfortable armchair and volume of speakers was set to a desirable level. They were advised to refrain from body and head movements and enjoy the music experience. Recording was divided into sessions of 30 minutes duration. The music streaming service was operated in radio mode, hence randomly selected songs (from the genre of their preference), were delivered to each participant while his/her brain activity was registered. Among the songs there were advertisements. That part of recordings was isolated from the rest. The recording procedure was integrated, in MATLAB, together with all necessary information from the streaming audio signal (i.e. song id, time stamps for the beginning and termination of each song). Participants evaluated each of the listened song, using as score one of the integers {1,2,3,4,5}, during a separate session just after the end of the recording. These scores, together with the associate patterns extracted from the recorded brain signals, comprised our training set. The overall procedure was repeated for every subject, in order to build an independent testing set.

### 4.4  Data Analysis

The preprocessing of signal included 50Hz component removal (by a built-in notch filter in *MuseIO* - the software that connects to and streams data from Muse), DC offset removal, and removal of the signal segments that corresponded to the 5 first second of each song (in order to avoid artificial transients).

The digital processing included i) band-pass filtering for deriving the brainwaves of standard brain rhythms, ii) Hilbert Transform for deriving their instantaneous amplitude and phase and iii) computation of the descriptors described in section 3.1. To increase frequency resolution, we divided the β rhythm into $β_{low}$ (13-20 Hz), and $β_{high}$ (20-30 Hz) sub-bands and derived descriptors separately. Similarly the γ rhythm was divided into $γ_{low}$ (30-49Hz) and $γ_{high}$ (51-90Hz). The recorded brain activity was segmented into overlapping windows of fixed duration (that during off-line experimentation had been varied between 30 and 100 seconds). The overall set of descriptors associated with the segments of brain activity had been used for selecting the best combination in a training-phase (applied collectively to all participants). A particular subset of descriptors (treated as a composite pattern including all the selected features) was extracted from each segment and utilized in an additional training phase (applied individually), during which an ELM-model was tailored to the user. In the testing phase, the subject specific ELM-model was applied to streaming composite patterns (representing segments of brain signals), in order to predict the listener's evaluation.

## 5   Results

### 5.1   Selecting Features - The synthesized biomarker

Feature selection was based on the Distance Correlation scores $\mathcal{R}$'s of all signal descriptors as averaged across all participants (Fig.2). The descriptors were ranked in descending order (regarding their music evaluation expressiveness) and a *dynamic programming* methodology was applied. Starting with the feature of highest $\mathcal{R}$, we traversed systematically the ranked list for combinations that would eventually maximize the Distance Correlation. This procedure led to a particular combination of 5 features, the **synthesized music appraisal biomarker**, which included the normalized temporal asymmetry index in $β_{low}$ band, relative energy of α band at temporal electrodes, $γ_{low} → γ_{high}$ PAC at FP1 sensor and relative energy of θ band at TP9. The relevance of this descriptor to music evaluation showed a dependence on the length of segment based on which it had been evaluated (using the timecourses from all participants, the averaged trace of Fig.3 was computed). It can be seen that the effectiveness of the biomarker constantly increases with the duration of the listened music. This kind of investigations can provide indications about how fast the automated music evaluation system could operate.

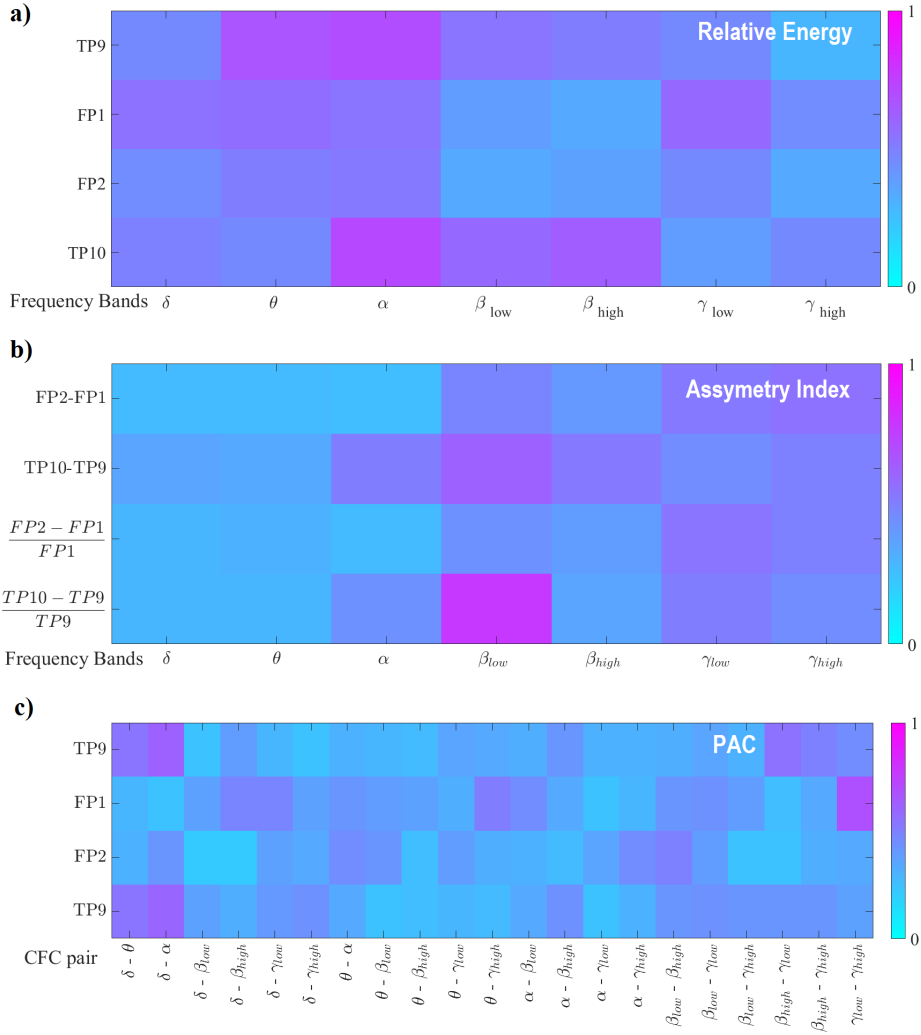

**Fig. 2.** Distance Correlation $\mathcal{R}$ of all employed signal descriptors.

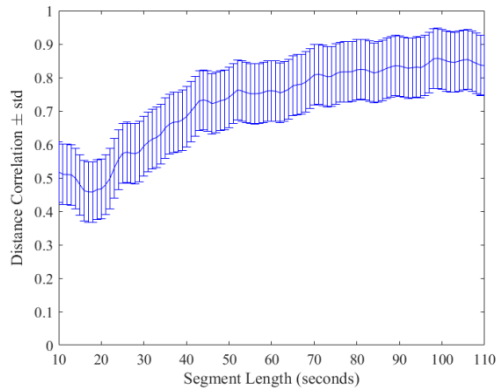

**Fig.3.** Distance Correlation $\mathcal{R}$ as a function of the segment length for the synthesized biomarker.

### 5.2 Designing the ELM

Using the common biomarker, an ELM was designed per listener. The number of neurons in the hidden player, the only parameter to be tuned in ELMs, was selected as the smallest number of neurons such that the training and testing error were converging at an acceptable level, lower than 0.01.

### 5.3 Evaluating the music-evaluation BCI

The proposed BCI was evaluated in two ways. During the first stage, the available data (the biomarker patterns of each of the five participants along with the associated ratings) were randomly partitioned in training and testing set, in a 60%-40% proportion for cross-validation. An ELM was trained using the former set and its performance was quantified using the latter set. The overall procedure was repeated 10 times, and averaged results are reported. The normalized root mean squared error (RMSE), as defined for regression tasks, was found to be 0.063±0.0093 (mean±std across participants). For comparison purposes, we also employed Support Vector Machines (SVMs), which performed slightly inferiorly. Although the difference was marginal, the very short training time was another factor in favor of employing ELMs.

During the second stage, a pilot online BCI was developed in order to realize the testing phase of the ELM in a realistic setting. Two of the previous subjects, participated in an additional recording session during which the already tailored ELMs were providing, based on segments of 90 seconds, a read-out of the subjective music evaluation. The predicted scores were registered and compared with the ones provided by the listeners just after the experimental session. The normalized RMSE was estimated to be 0.09 and 0.07.

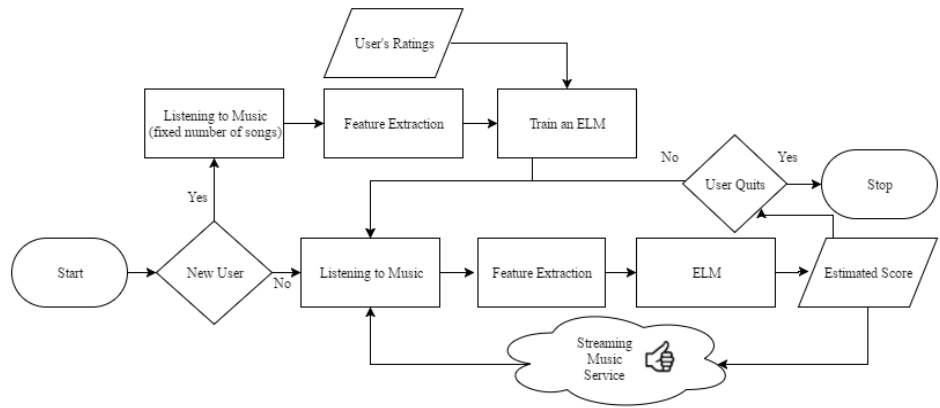

**Fig. 4.** Flow chart of the proposed music evaluation BCI

## 6 Discussion

This paper constitutes the report from a pilot study, during which we attempted to associate the listener's brainwaves with the subjective aesthetic pleasure induced by music. The main novelty of the presented proof-of-concept, is that it was realized based on a modern consumer EEG device and in conjunction with a popular on-demand music streaming service (Spotify). Our results indicate that signals from a restricted number of sensors (located over frontal and temporal brain areas) can be combined in a computable biomarker reflecting the listener's subjective music evaluation. This brainwaves' derivative can be efficiently decoded by regression-ELM and therefore leads to a reliable readout from the listener. The main advantage of the approach is that complies with idea of employing EEG-wearables in daily activities and is readily embedded within the contemporary on-demand music streaming services (see Fig.4). However, the problem of artifacts (noisy signals of biological origin) has not been addressed yet. For the presented results, the participants had been asked to limit body/head movements and facial expressions and as much as possible. Hence, before employing such a system to naturalistic recordings, methodologies for real-time artifact suppression (as in [27]) have to be incorporated.

Today, we live in the world of *Internet of Things* (IoT) where the interconnectedness among devices has already been anticipated and supportive technologies are now being realized [28]. However, to achieve a seamless integration of technology in people's lives, there is still much room for improvement. In a world beyond IoT, technology would proactively facilitate people's expectations and wearable devices would transparently interface with systems and services. To enable such scenarios would require to move from the IoT to the *Internet of People* (IoP) [29]. People would then participate as first-class citizens relishing the benefits of holistic human-friendly applications. As such, we present a pragmatic use of a consumer BCI that ideally fits in the anticipated forms of the digital music universe and demonstrate a first series of encouraging experimental results.